\documentclass{article}

\usepackage{PRIMEarxiv}

\usepackage[utf8]{inputenc} 
\usepackage[T1]{fontenc}    
\usepackage{hyperref}       
\usepackage{url}            
\usepackage{booktabs}       
\usepackage{amsfonts}       
\usepackage{nicefrac}       
\usepackage{microtype}      
\usepackage{amsmath}        
\usepackage{fancyhdr}       
\usepackage{graphicx}       
\graphicspath{{media/}}     

\usepackage{caption}
\usepackage{makecell}
\usepackage{longtable}      
\usepackage{tabularx}       
\usepackage{adjustbox}      
\usepackage{rotating}       
\usepackage[frozencache,cachedir=minted-cache]{minted}
\hypersetup{hidelinks}

\title{Beyond Benchmarks: The Economics of AI Inference
}

\author{
  WiNGPT Team\thanks{\textbf{Authors are listed in section \ref{sec:authors}}} \\
  \\
  Winning Health AI Research \\
  \texttt{Correspondence to wair@winning.com.cn} \\
}

\begin{document}
\maketitle

\begin{abstract}
The inference cost of Large Language Models (LLMs) has become a critical factor in determining their commercial viability and widespread adoption. This paper introduces a quantitative ``economics of inference'' framework, treating the LLM inference process as a compute-driven intelligent production activity. We analyze its marginal cost, economies of scale, and quality of output under various performance configurations. Based on empirical data from WiNEval-3.0, we construct the first ``LLM Inference Production Frontier,'' revealing three principles: diminishing marginal cost, diminishing returns to scale, and an optimal cost-effectiveness zone. This paper not only provides an economic basis for model deployment decisions but also lays an empirical foundation for the future market-based pricing and optimization of AI inference resources.
\end{abstract}

\section{The Cost Challenge in the Real World}

As Large Language Models (LLMs) \cite{brown2020language,touvron2023llama} rapidly expand into healthcare, scientific research, and industrial applications, the economic cost of the inference stage has become the primary bottleneck limiting their scalability.

Most existing research focuses on single dimensions like model accuracy or inference speed \cite{zheng2023judging}, overlooking the equally critical cost constraints in real-world business environments. Models that excel on academic leaderboards often become unfeasible for large-scale commercial deployment due to excessive per-unit compute costs \cite{patterson2021carbon}. This presents an ``impossible trinity'': Model Quality (Q) -- Inference Performance (P) -- Economic Cost (C). Any model's engineering deployment must strike a balance among these three.

This paper proposes a systematic framework for quantifying inference costs. Based on experimental data from real-world business workloads, we map out the ``cost-quality Pareto frontier'' for models to guide the selection of models and resources for different task scenarios. Using this framework, we can:

\begin{enumerate}
    \item More accurately estimate and plan the scale of GPU procurement.
    \item Select suitable models for specific tasks, balancing cost, performance, and quality.
    \item Optimize inference concurrency and scheduling strategies, and guide corporate technology roadmaps.
\end{enumerate}

\section{The Production Function of Intelligence}

From an economic perspective, the LLM inference process can be formalized as an ``intelligent production activity.'' We simplify this process into a production function:

\begin{equation}
\text{Intelligence} = f(\text{Cost}, \text{Model})
\end{equation}

This function reveals an economic trade-off: achieving a higher level of intelligence typically requires a higher economic cost \cite{kaplan2020scaling,hoffmann2022training}. This cost may stem from using more complex, larger-parameter, or more advanced models, larger-scale computing power, or longer processing times. An ideal model can produce higher-quality content at a lower cost, defining the market's efficiency benchmark---the highest level of intelligence achievable at a specific cost.

Let's proceed step-by-step, starting with defining the cost.

\section{Estimating Hourly GPU Cost}

In a typical bare-metal server deployment, the hourly cost of a single GPU consists of three components: depreciation, power consumption, and maintenance \cite{singla2022understanding}.

\begin{equation}
\text{Hourly GPU Cost} \approx \text{Depreciation} + \text{Power Consumption} + \text{Maintenance}
\end{equation}

We estimate this cost by considering the GPU purchase price ($P$), depreciation period ($Y$), utilization rate ($u$), average power consumption (kW), data center Power Usage Effectiveness (PUE), electricity price ($E$), and annual maintenance fee rate ($m$). For detailed calculation formulas and parameter values, please refer to Appendix~\ref{app:appendix-a}.

Taking the A800 80G as an example, under common assumptions, its baseline hourly cost per card is approximately \$0.79/hour, generally falling within the \$0.51--\$0.99/hour range.

On the other hand, major cloud platforms offer comparable computing power at \$2.82--\$5.64/hour. For example, AWS P4de instances are priced at \$5.08/hour, and Alibaba Cloud's gn7e-c16g1.4xlarge is \$4.80/hour (including electricity and maintenance).

Companies often weigh the trade-offs between ``building a self-hosted GPU cluster'' and ``renting from the cloud.'' Cloud solutions offer more flexibility for small to medium-scale or fluctuating workloads, while self-hosted clusters provide better marginal cost advantages for sustained high-load scenarios. This paper uses \$0.79/hour as a neutral baseline to ensure comparability across models.

\section{Estimating Inference Cost}

Calculating inference cost essentially involves translating task execution time into hardware cost. In this framework, we calculate the total cost to complete the entire WiNEval-3.0 test set (containing 2,993 requests).

Core formula:
\begin{equation}
\text{Total Test Set Cost} = \text{Total Hourly GPU Cost} \times \text{Total Execution Time}
\end{equation}

In this test, our environment and cost baseline are:

\begin{itemize}
    \item Environment: A800 80G $\times$ 2 cards
    \item Dual-card hourly cost: \$1.58
\end{itemize}

Therefore, for a test task with a total duration of $T$ seconds, the total cost is calculated as:
\begin{equation}
\text{Total Test Set Cost (\$)} = 1.58 \times \frac{T}{3600}
\end{equation}

This formula will be used directly for all subsequent cost conversions for different models and concurrency configurations.

\begin{quote}
As a professional evaluation set for the medical field, WiNEval's tasks are derived from real clinical applications, covering 10 core scenarios such as medical licensing exams, clinical diagnosis, quality control, and medical text correction. We choose WiNEval-3.0 as our benchmark not only because it covers multidimensional medical tasks but also because its task structure exhibits ``representative economic load characteristics'':

\begin{enumerate}
    \item The length distribution among tasks approximates the ``long-tail distribution'' of real-world applications, reflecting resource utilization fluctuations during inference.
    \item The input and output sizes are stable, allowing for the quantification of per-task cost.
    \item Tasks can be executed concurrently and independently, making it suitable for constructing an ``inference production function'' (i.e., the curve of task throughput versus concurrency growth).
    \item The costs here are estimates based on performance data, reflecting the relative cost differences under various concurrency levels.
\end{enumerate}
\end{quote}

\section{Evaluation Dimensions and Optimization Goals}

To make scientific model selection decisions, we must evaluate models comprehensively across three dimensions: performance, quality, and cost.

\subsection{Performance}

Performance metrics focus on the system's operational efficiency. We select three core metrics for a comprehensive assessment:

\begin{table}[htbp]
\centering
\small
\begin{tabularx}{\textwidth}{lXXl}
\toprule
\textbf{Metric} & \textbf{Meaning} & \textbf{Business Significance} & \textbf{Target} \\
\midrule
Total Completion Time (s) & Total time to complete all requests & Measures system throughput \& efficiency & As short as possible \\
Avg. TTFT (s) & Average time from request to receiving the first token & Directly impacts user interaction experience & $<$ 1s \\
Avg. Throughput (tokens/s) & Number of tokens generated per second & Directly impacts user interaction experience & $>$ 20 tok/s \\
\bottomrule
\end{tabularx}
\caption{Performance Metrics and Optimization Targets}
\end{table}

The performance baselines above are examples set for scenarios like interactive clinical decision support, where TTFT $<$ 1s and average throughput $>$ 20 tokens/s are crucial for ensuring smooth product interaction. In other medical scenarios, such as batch medical record summarization or offline medical literature analysis, the business focus may shift to total completion time and throughput, with less stringent latency requirements. Performance thresholds can be dynamically adjusted based on specific business needs.

\subsection{Quality}

No matter how fast a model runs or how low its cost is, it is meaningless if its generated content fails to meet the stringent requirements (or accuracy) of clinical applications. This framework introduces the WiNEval-3.0 average score as the core quality metric to measure a model's comprehensive abilities in medical knowledge understanding, clinical reasoning, and instruction following.

\subsection{Cost}

Cost is the ultimate benchmark that determines whether a technical solution can be scaled. We use the total test set cost derived in Sections~3 and 4 as the core unit cost metric, which unifies hardware, energy, and time into quantifiable financial data. The ``cost'' here is not a variable independent of performance but a relative computational expense calculated based on performance thresholds, representing the economic efficiency of resource utilization. To find the optimal performance configuration for each model, we recorded their core metrics under different concurrency pressures in a unified test environment. For detailed data, please refer to Appendix~\ref{app:appendix-b}.

\section{The Relationship Between Performance, Cost, and Quality}

\subsection{The Balance Point of Performance and Cost}

Based on the data from Appendix~\ref{app:appendix-b}, we have plotted a 3D AI inference production frontier graph showing model quality versus inference cost at optimal performance configurations. The size of the bubbles represents the model's parameter count.

\begin{figure}[htbp]
\centering
\includegraphics[width=0.8\textwidth]{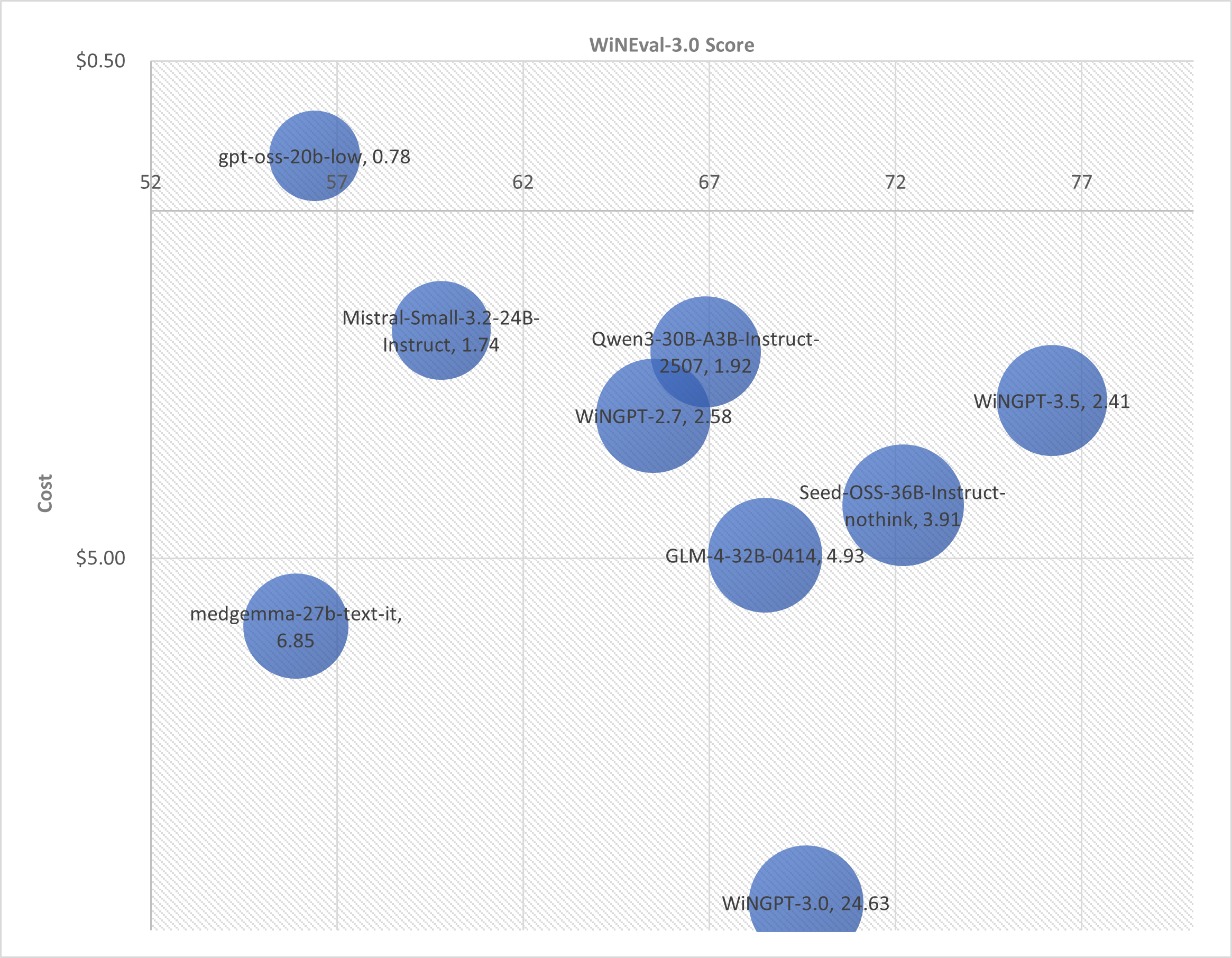}
\caption{Model Quality vs. Inference Cost - 3D Pareto Frontier}
\end{figure}

As shown in the figure, we can quickly identify the ``high-value'' models located in the upper-left corner (low cost, high quality) and spot the ``outliers'' with extreme cost or quality performance. Taking the test data for WiNGPT-3.5 as an example, we found that:

\begin{enumerate}
    \item \emph{Increasing concurrency reduces total time}: As concurrency increases from 8 to 48, the total completion time drops from 2034 seconds to 774 seconds. This indicates that before GPU compute power is saturated, increasing concurrency is the most effective way to amortize fixed overhead and reduce per-unit time costs.
    \item \emph{A performance inflection point exists}: When concurrency is further increased from 48, throughput drops sharply, and TTFT may also increase dramatically. Nearly all models have an optimal concurrency range. Beyond this range, system overhead soars, service quality declines, and the marginal cost-benefit diminishes or even becomes negative.
    \item \emph{Determining the optimal performance configuration}: Our goal is to find the concurrency setting with the lowest cost (i.e., shortest total completion time) while meeting the performance baselines (e.g., throughput $>$ 20 tokens/s and latency $<$ 1s). For WiNGPT-3.5, a concurrency of 48 is its optimal inference configuration under this test.
\end{enumerate}

\subsection{Balancing Model Performance, Quality, and Cost}

Finally, we selected the optimal performance configuration for each model and conducted a horizontal comparison. We have specifically included the input/output and total token counts to help explain the cost differences for some models.

\begin{table}[htbp]
\centering
\caption{Model Performance, Cost, and Quality Comparison at Optimal Configurations}
\begin{adjustbox}{max width=\textwidth}
\footnotesize
\begin{tabular}{lcccccccccc}
\toprule
\textbf{Model} & \textbf{Params} & \textbf{Conc.} & \textbf{Total Time} & \textbf{Avg. TTFT} & \textbf{Input} & \textbf{Output} & \textbf{Total} & \textbf{Throughput} & \textbf{Cost} & \textbf{Score} \\
 & \textbf{(B)} & & \textbf{(s)} & \textbf{(s)} & \textbf{Tokens} & \textbf{Tokens} & \textbf{Tokens} & \textbf{(tok/s)} & \textbf{(\$)} & \\
\midrule
WiNGPT-3.5 & 30 & 48 & 774.11 & 0.147 & 1,347,535 & 796,836 & 2,144,371 & 21.45 & 0.34 & \textbf{76.2} \\
Seed-OSS-36B & 36 & 16 & 1255.4 & 0.222 & 1,238,191 & 513,012 & 1,751,203 & 25.54 & 0.55 & 72.2 \\
WiNGPT-3.0 & 32 & 16 & 7916.62 & 0.142 & 1,347,535 & 3,440,393 & 4,787,928 & 27.16 & 3.47 & 69.6 \\
GLM-4-32B & 32 & 8 & 1583.45 & 0.119 & 1,226,578 & 415,190 & 1,641,768 & 32.78 & 0.69 & 68.5 \\
Qwen3-30B & 30 & 64 & 616.77 & 0.168 & 1,347,535 & 790,773 & 2,138,308 & 20.03 & 0.27 & 66.9 \\
WiNGPT-2.7 & 32 & 16 & 830.37 & 0.156 & 1,347,535 & 345,306 & 1,692,841 & 25.99 & 0.36 & 65.5 \\
Mistral-Small & 24 & 64 & 559.2 & 0.276 & 2,113,182 & 813,781 & 2,926,963 & 22.74 & 0.25 & 59.8 \\
medgemma-27b & 27 & 32 & 2200.62 & 0.190 & 1,399,753 & 1,411,583 & 2,811,336 & 20.05 & 0.97 & 55.9 \\
gpt-oss-20b & 20 & 64 & 249.17 & 0.073 & 1,495,464 & 398,699 & 1,894,163 & 25.00 & \textbf{0.11} & 56.4 \\
\bottomrule
\end{tabular}
\end{adjustbox}
\end{table}

Looking at the table, most models fall within the ``sweet spot'' of under \$1.40 in cost. WiNGPT-3.5 (76.2 score, \$0.34) is the overall leader, providing the highest quality at a highly competitive cost, making it the best choice for balancing effectiveness and budget. It is followed by Seed-OSS-36B (72.2 score, \$0.55), which is also in the high-quality range but at a higher cost and slightly lower efficiency.

At the extremes of the cost distribution, there are also ``extreme options'' worth noting. gpt-oss-20b-low, with a cost of only \$0.11, is a ``potential contender'' for cost-effectiveness. Mistral-Small controls its final cost well (\$0.25), but its input token count (2.11 million) is much higher than most models (around 1.3 million), indicating its tokenizer is less efficient for Chinese, requiring more tokens to process the same text.

The other extreme is WiNGPT-3.0, whose high cost of \$3.47 makes it an ``outlier.'' The root of its cost lies in its massive generation volume---its total output tokens are 4 to 8 times that of most models. This reveals its true identity: a ``thinking'' model built for complex reasoning. Its output includes detailed chains of thought, making it unsuitable for routine medical conversations but a specialized tool for professional domains requiring process transparency and logical traceability, such as complex case analysis or drafting treatment plans. Its high cost is not a flaw but a direct reflection of its deep reasoning capabilities.

\section{Limitations}

Although this framework provides a systematic evaluation method, its limitations must be acknowledged:

\begin{enumerate}
    \item \emph{Training costs are not included}: This framework focuses on the inference deployment stage and does not cover the costs associated with model fine-tuning or continuous training, which are important components in customized applications.
    \item \emph{Dependency on a specific software/hardware stack}: The evaluation results are based on specific hardware and inference services \cite{kwon2023efficient}. Changing the GPU, inference engine, or quantization strategy could significantly alter the performance and cost data.
    \item \emph{Proxy nature of benchmark scores}: While WiNEval-3.0 serves as a high-quality proxy metric, it is not entirely equivalent to a model's final performance in specific, specialized clinical business scenarios.
    \item \emph{Lack of statistical confidence analysis}: Future work should introduce confidence intervals and sensitivity analysis to verify the robustness of the results.
    \item \emph{Upfront capital expenditure is not considered}: The massive investment required to purchase tens or even hundreds of GPUs at once. This decision-making barrier directly influences the choice of technology roadmap (e.g., self-hosting vs.\ cloud rental) and may render some theoretically optimal solutions (which rely on expensive hardware) infeasible in reality. The final decision must still be made based on a comprehensive assessment of factors like real-world business volume.
\end{enumerate}

\section{Conclusion}

This paper, based on empirical data from WiNEval-3.0, has constructed and validated a data-driven LLM selection framework that integrates performance, cost, and quality. It moves away from abstractly ranking models in experimental settings and instead examines these three core business variables under real-world business loads. The core value of this framework lies in:

\begin{enumerate}
    \item \emph{Being rooted in real-world tasks}: All conclusions are derived from stress tests in clinical medical scenarios, ensuring that the evaluation results have direct guiding significance for production environments.
    \item \emph{High portability}: By adjusting core parameters like hourly GPU cost, the framework can be easily adapted to different hardware infrastructures or cloud service platforms.
    \item \emph{Providing a quantifiable basis for decision-making}: It enables a shift from ``gut feeling'' to ``data'' for critical corporate decisions regarding GPU investment, model selection, and concurrency optimization.
\end{enumerate}

Ultimately, our analysis reveals that there is no one-size-fits-all ``best model.'' Instead, there is a diverse ecosystem of models, each achieving its optimal cost-effectiveness at a specific concurrency configuration. This framework provides the first quantifiable decision-making tool for selecting the best AI technology within a limited budget, marking a key shift in the industry's focus: from the pursuit of endless model parameters to the efficiency of engineered, measurable application deployment.

\section{Authors}
\label{sec:authors}
Boqin Zhuang, Jiacheng Qiao, Mingqian Liu, Mingxing Yu, Ping Hong, Rui Li, Xiaoxia Song, Xiangjun Xu, Xu Chen, Yaoyao Ma, Yujie Gao

\bibliographystyle{unsrt}
\bibliography{main}

\begin{thebibliography}{1}

\bibitem{brown2020language}
Tom Brown, Benjamin Mann, Nick Ryder, Melanie Subbiah, Jared~D Kaplan, Prafulla Dhariwal, Arvind Neelakantan, Pranav Shyam, Girish Sastry, Amanda Askell, et~al.
\newblock Language models are few-shot learners.
\newblock {\em Advances in neural information processing systems}, 33:1877--1901, 2020.

\bibitem{touvron2023llama}
Hugo Touvron, Louis Martin, Kevin Stone, Peter Albert, Amjad Almahairi, Yasmine Babaei, Nikolay Bashlykov, Soumya Batra, Prajjwal Bhargava, Shruti Bhosale, et~al.
\newblock Llama 2: Open foundation and fine-tuned chat models.
\newblock {\em arXiv preprint arXiv:2307.09288}, 2023.

\bibitem{zheng2023judging}
Lianmin Zheng, Wei-Lin Chiang, Ying Sheng, Siyuan Zhuang, Zhanghao Wu, Yonghao Zhuang, Zi~Lin, Zhuohan Li, Dacheng Li, Eric Xing, et~al.
\newblock Judging llm-as-a-judge with mt-bench and chatbot arena.
\newblock {\em Advances in Neural Information Processing Systems}, 36, 2023.

\bibitem{patterson2021carbon}
David Patterson, Joseph Gonzalez, Quoc Le, Chen Liang, Lluis-Miquel Munguia, Daniel Rothchild, David So, Maud Texier, and Jeff Dean.
\newblock Carbon emissions and large neural network training.
\newblock {\em arXiv preprint arXiv:2104.10350}, 2021.

\bibitem{kaplan2020scaling}
Jared Kaplan, Sam McCandlish, Tom Henighan, Tom~B Brown, Benjamin Chess, Rewon Child, Scott Gray, Alec Radford, Jeffrey Wu, and Dario Amodei.
\newblock Scaling laws for neural language models.
\newblock {\em arXiv preprint arXiv:2001.08361}, 2020.

\bibitem{hoffmann2022training}
Jordan Hoffmann, Sebastian Borgeaud, Arthur Mensch, Elena Buchatskaya, Trevor Cai, Eliza Rutherford, Diego de~Las Casas, Lisa~Anne Hendricks, Johannes Welbl, Aidan Clark, et~al.
\newblock Training compute-optimal large language models.
\newblock {\em arXiv preprint arXiv:2203.15556}, 2022.

\bibitem{singla2022understanding}
Advith Singla, Rajarshi~Roy Choudhury, Hanchen Zhao, and Alexey Tumanov.
\newblock Understanding gpu memory growth: A visualization approach.
\newblock In {\em Proceedings of the 14th ACM Workshop on Hot Topics in Storage and File Systems}, pages 86--93, 2022.

\bibitem{kwon2023efficient}
Woosuk Kwon, Zhuohan Li, Siyuan Zhuang, Ying Sheng, Lianmin Zheng, Cody~Hao Yu, Joseph Gonzalez, Hao Zhang, and Ion Stoica.
\newblock Efficient memory management for large language model serving with pagedattention.
\newblock In {\em Proceedings of the 29th Symposium on Operating Systems Principles}, pages 611--626, 2023.

\end{thebibliography}

\newpage

\appendix
\section*{Appendix}

\section{GPU Hourly Cost Estimation Method}
\label{app:appendix-a}

We use the following formulas for estimation:

\begin{align}
\text{Depreciation} &= \frac{P}{Y \times 8760 \times u} \\
\text{Power Consumption} &= kW \times PUE \times E \\
\text{Maintenance} &= \frac{P \times m}{8760}
\end{align}

The parameters in the formulas are defined with reference values as shown in the table below:

\begin{table}[htbp]
\centering
\caption{GPU Cost Calculation Parameters}
\begin{tabular}{lll}
\toprule
\textbf{Parameter} & \textbf{Meaning} & \textbf{Reference Value} \\
\midrule
P & GPU Purchase Price (\$) & \textit{Specific value} \\
Y & Depreciation Period (years) & 3 \\
u & GPU Utilization Rate (1 for 100\%) & 1 (baseline) \\
kW & Average Power Consumption (kilowatts) & \textit{Specific value} \\
PUE & Data Center Power Usage Effectiveness & 1.3 $\sim$ 1.5 \\
E & Electricity Price (CNY/kWh) & \textit{Specific value} \\
m & Annual Maintenance Fee Rate & 3\% \\
\bottomrule
\end{tabular}
\end{table}

\begin{quote}
\textbf{Note on GPU Utilization Rate ($u$):}

Here, $u$ is set to 1 (100\%) as an idealized theoretical baseline to calculate the opportunity cost of the hardware. It represents the base hourly cost of the hardware operating at full capacity around the clock. In practice, the average GPU utilization is often below 100\%, for example, 70\% ($u=0.7$). In that case, the cost per effective compute hour would increase (Effective Hourly Cost = Base Hourly Cost / $u$). This framework uses $u=1$ as a standardized baseline for fair comparison across different hardware and workloads.
\end{quote}

In a real-world environment, purchase price, power consumption, electricity rates, PUE, and utilization rates all significantly impact cost. We estimate based on common assumptions:

\begin{itemize}
    \item Depreciation: 120,000 / (3 $\times$ 8760) / 7.09 $\approx$ \$0.64/hour
    \item Power Consumption: 0.4 $\times$ 1.5 $\times$ 1.0 / 7.09 $\approx$ \$0.08/hour
    \item Maintenance: 120,000 $\times$ 0.03 / 8760 / 7.09 $\approx$ \$0.06/hour
    \item Total cost $\approx$ 0.64 + 0.08 + 0.06 $\approx$ \$0.78/hour
\end{itemize}

\textbf{Conclusion:}

\begin{itemize}
    \item The hourly cost of an A800 80G is approximately \$0.79/hour under common assumptions.
    \item Depending on the purchase price range, the cost fluctuates between \$0.51--\$0.99/hour.
    \item Costs will further increase with higher power consumption, electricity rates, or lower utilization.
    \item This calculation does not include additional costs such as the host server, rack, networking, or labor, reflecting only the GPU's bare hardware cost range.
\end{itemize}

\section{Performance, Cost, and Quality Data for Each Model Under Different Concurrency Pressures}
\label{app:appendix-b}

{\fontsize{7}{9}\selectfont
\begin{longtable}{lccccccccc}
\caption{Detailed Performance Data Across Different Concurrency Levels} \\
\toprule
\textbf{Model / Conc.} & \textbf{Req.} & \textbf{Time} & \textbf{TTFT} & \textbf{Input} & \textbf{Output} & \textbf{Total} & \textbf{Tput} & \textbf{Cost} \\
 & & \textbf{(s)} & \textbf{(s)} & \textbf{Tok.} & \textbf{Tok.} & \textbf{Tok.} & \textbf{(tok/s)} & \textbf{(\$)} \\
\midrule
\endfirsthead

\toprule
\textbf{Model / Conc.} & \textbf{Req.} & \textbf{Time} & \textbf{TTFT} & \textbf{Input} & \textbf{Output} & \textbf{Total} & \textbf{Tput} & \textbf{Cost} \\
 & & \textbf{(s)} & \textbf{(s)} & \textbf{Tok.} & \textbf{Tok.} & \textbf{Tok.} & \textbf{(tok/s)} & \textbf{(\$)} \\
\midrule
\endhead

\endfoot

\bottomrule
\endlastfoot

\textbf{WiNGPT-2.7} & & & & & & & & \\
8 & 2993 & 1386.53 & 0.119 & 1,347,535 & 344,068 & 1,691,603 & 31.02 & 0.61 \\
\textbf{16} & \textbf{2993} & \textbf{830.37} & \textbf{0.156} & \textbf{1,347,535} & \textbf{345,306} & \textbf{1,692,841} & \textbf{25.99} & \textbf{0.36} \\
32 & 2993 & 561.57 & 0.224 & 1,347,535 & 356,912 & 1,704,447 & 19.86 & 0.25 \\
48 & 2993 & 461.3 & 0.295 & 1,347,535 & 344,281 & 1,691,816 & 15.55 & 0.20 \\
64 & 2993 & 422.36 & 0.353 & 1,347,535 & 345,064 & 1,692,599 & 12.77 & 0.19 \\
128 & 2993 & 378.29 & 0.614 & 1,347,535 & 346,518 & 1,694,053 & 7.16 & 0.17 \\
\textbf{GLM-4-32B-0414} & & & & & & & & \\
\textbf{8} & \textbf{2993} & \textbf{1583.45} & \textbf{0.119} & \textbf{1,226,578} & \textbf{415,190} & \textbf{1,641,768} & \textbf{32.78} & \textbf{0.69} \\
16 & 2993 & 1694.2 & 0.165 & 1,226,578 & 448,842 & 1,675,420 & 16.56 & 0.74 \\
32 & 2993 & 1268.64 & 0.231 & 1,226,578 & 446,399 & 1,672,977 & 16.56 & 0.56 \\
48 & 2993 & 475.83 & 0.303 & 1,226,578 & 419,237 & 1,645,815 & 16.56 & 0.21 \\
64 & 2993 & 422.93 & 0.373 & 1,226,578 & 410,913 & 1,637,491 & 16.56 & 0.19 \\
128 & 2993 & 420.43 & 0.877 & 1,226,578 & 424,935 & 1,651,513 & 16.56 & 0.18 \\
\textbf{gpt-oss-20b-low} & & & & & & & & \\
8 & 2993 & 781.44 & 0.054 & 1,495,464 & 398,621 & 1,894,085 & 63.76 & 0.34 \\
16 & 2993 & 585.78 & 0.042 & 1,495,464 & 429,743 & 1,925,207 & 45.85 & 0.26 \\
32 & 2993 & 331.56 & 0.048 & 1,495,464 & 396,885 & 1,892,349 & 37.41 & 0.15 \\
48 & 2993 & 259.28 & 0.059 & 1,495,464 & 395,878 & 1,891,342 & 31.81 & 0.11 \\
\textbf{64} & \textbf{2993} & \textbf{249.17} & \textbf{0.073} & \textbf{1,495,464} & \textbf{398,699} & \textbf{1,894,163} & \textbf{25.00} & \textbf{0.11} \\
128 & 2993 & 164.17 & 0.230 & 1,495,464 & 388,587 & 1,884,051 & 18.49 & 0.07 \\
\textbf{WiNGPT-3.0} & & & & & & & & \\
8 & 2993 & 13593.47 & 0.123 & 1,347,535 & 3,518,378 & 4,865,913 & 32.35 & 5.96 \\
\textbf{16} & \textbf{2993} & \textbf{7916.62} & \textbf{0.142} & \textbf{1,347,535} & \textbf{3,440,393} & \textbf{4,787,928} & \textbf{27.16} & \textbf{3.47} \\
32 & 2993 & 6252.95 & 0.191 & 1,347,535 & 3,517,540 & 4,865,075 & 17.58 & 2.74 \\
48 & 2993 & 5925.54 & 0.230 & 1,347,535 & 3,636,012 & 4,983,547 & 12.78 & 2.60 \\
64 & 2993 & 5305.93 & 5.219 & 1,347,535 & 3,736,644 & 5,084,179 & 11.00 & 2.33 \\
128 & 2993 & 4736.77 & 57.404 & 1,347,535 & 3,841,930 & 5,189,465 & 6.34 & 2.08 \\
\textbf{Seed-OSS-36B} & & & & & & & & \\
8 & 2993 & 2134.78 & 0.168 & 1,238,191 & 509,206 & 1,747,397 & 29.82 & 0.94 \\
\textbf{16} & \textbf{2993} & \textbf{1255.4} & \textbf{0.222} & \textbf{1,238,191} & \textbf{513,012} & \textbf{1,751,203} & \textbf{25.54} & \textbf{0.55} \\
32 & 2993 & 1792.73 & 0.337 & 1,238,191 & 708,137 & 1,946,328 & 12.34 & 0.79 \\
48 & 2993 & 671.64 & 0.410 & 1,238,191 & 506,833 & 1,745,024 & 15.72 & 0.29 \\
64 & 2993 & 629.68 & 0.555 & 1,238,191 & 507,281 & 1,745,472 & 12.59 & 0.28 \\
128 & 2993 & 578.66 & 1.195 & 1,238,191 & 507,014 & 1,745,205 & 6.85 & 0.25 \\
\textbf{medgemma-27b} & & & & & & & & \\
8 & 2993 & 5371.46 & 0.109 & 1,399,753 & 1,421,097 & 2,820,850 & 33.07 & 2.36 \\
16 & 2993 & 3706.2 & 0.133 & 1,399,753 & 1,618,060 & 3,017,813 & 27.29 & 1.63 \\
\textbf{32} & \textbf{2993} & \textbf{2200.62} & \textbf{0.190} & \textbf{1,399,753} & \textbf{1,411,583} & \textbf{2,811,336} & \textbf{20.05} & \textbf{0.97} \\
48 & 2993 & 2056.75 & 0.219 & 1,399,753 & 1,520,875 & 2,920,628 & 15.41 & 0.90 \\
64 & 2993 & 2006.41 & 0.263 & 1,399,753 & 1,498,920 & 2,898,673 & 11.67 & 0.88 \\
128 & 2993 & 1733.44 & 1.201 & 1,399,753 & 1,418,759 & 2,818,512 & 6.39 & 0.76 \\
\textbf{Mistral-Small} & & & & & & & & \\
8 & 2993 & 1938.63 & 0.108 & 2,113,182 & 811,852 & 2,925,034 & 52.35 & 0.85 \\
16 & 2993 & 1117 & 0.132 & 2,113,182 & 810,838 & 2,924,020 & 45.37 & 0.49 \\
32 & 2993 & 738.22 & 0.173 & 2,113,182 & 824,301 & 2,937,483 & 34.89 & 0.32 \\
48 & 2993 & 630.79 & 0.224 & 2,113,182 & 811,290 & 2,924,472 & 26.79 & 0.28 \\
\textbf{64} & \textbf{2993} & \textbf{559.2} & \textbf{0.276} & \textbf{2,113,182} & \textbf{813,781} & \textbf{2,926,963} & \textbf{22.74} & \textbf{0.25} \\
128 & 2993 & 456.82 & 0.539 & 2,113,182 & 813,335 & 2,926,517 & 13.91 & 0.20 \\
\textbf{Qwen3-30B} & & & & & & & & \\
8 & 2993 & 1381.05 & 0.067 & 1,347,535 & 783,226 & 2,130,761 & 70.89 & 0.61 \\
16 & 2993 & 1059.78 & 0.093 & 1,347,535 & 835,127 & 2,182,662 & 49.25 & 0.47 \\
32 & 2993 & 1114.56 & 0.123 & 1,347,535 & 965,555 & 2,313,090 & 27.07 & 0.49 \\
48 & 2993 & 739.24 & 0.141 & 1,347,535 & 848,943 & 2,196,478 & 23.93 & 0.32 \\
\textbf{64} & \textbf{2993} & \textbf{616.77} & \textbf{0.168} & \textbf{1,347,535} & \textbf{790,773} & \textbf{2,138,308} & \textbf{20.03} & \textbf{0.27} \\
128 & 2993 & 336.6 & 0.450 & 1,347,535 & 782,911 & 2,130,446 & 18.17 & 0.15 \\
\textbf{WiNGPT-3.5} & & & & & & & & \\
8 & 2993 & 2034.05 & 0.103 & 1,347,535 & 932,262 & 2,279,797 & 57.29 & 0.89 \\
16 & 2993 & 1098.77 & 0.117 & 1,347,535 & 762,906 & 2,110,441 & 43.40 & 0.48 \\
32 & 2993 & 863.7 & 0.134 & 1,347,535 & 773,120 & 2,120,655 & 27.97 & 0.38 \\
\textbf{48} & \textbf{2993} & \textbf{774.11} & \textbf{0.147} & \textbf{1,347,535} & \textbf{796,836} & \textbf{2,144,371} & \textbf{21.45} & \textbf{0.34} \\
64 & 2993 & 599.03 & 0.163 & 1,347,535 & 714,003 & 2,061,538 & 18.62 & 0.26 \\
128 & 2993 & 668.04 & 0.319 & 1,347,535 & 813,350 & 2,160,885 & 9.51 & 0.29 \\

\end{longtable}
}  

\begin{quote}
\textbf{Our concurrency parameter design follows this logic:}

\begin{itemize}
    \item With actual GPU utilization as the primary observation metric, the concurrency range from 8 to 64 covers three typical workloads: low load, medium load, and saturated load.
    \item The data in the table may have slight fluctuations. This is mainly due to two factors: first, model generation has some inherent randomness; second, the dynamic batching and scheduling mechanisms of inference frameworks like vLLM introduce execution variations. For example, although WiNGPT-3.5 and Qwen3-30B-A3B-Instruct-2507 have the same architecture, their different post-training strategies lead to variations in the total number of tokens generated for the same task set, which can also cause deviations in their average throughput.
\end{itemize}
\end{quote}

\end{document}